\RequirePackage{fix-cm}
\documentclass[twocolumn]{svjour3}          
\smartqed  
\usepackage{graphicx}

\usepackage{amsmath}
\usepackage{amssymb}
\usepackage{amsfonts}
\usepackage{booktabs}
\usepackage{multirow}
\usepackage{subfig}
\usepackage{subfloat}
\usepackage{floatflt}
\usepackage{wrapfig}
\usepackage[linesnumbered, ruled,vlined]{algorithm2e}
\SetKwRepeat{Do}{do}{while}%
\usepackage{url}
\usepackage{xcolor}
\usepackage{adjustbox}
\usepackage{threeparttable}
\usepackage{lscape}
\usepackage{pdflscape}
\usepackage{moresize}
\usepackage{listings}
\usepackage{cprotect}

\def\vector#1{\mbox{\boldmath $#1$}}
\newcommand{\CR}{C}

\begin{document}

\title{Review and Analysis of Three Components of Differential Evolution Mutation Operator in MOEA/D-DE}



\author{Ryoji Tanabe \and Hisao Ishibuchi}



  
\institute{R. Tanabe and H. Ishibuchi \at
Shenzhen Key Laboratory of Computational Intelligence, University Key Laboratory of Evolving Intelligent Systems of Guangdong Province, Department of Computer Science and Engineering, Southern University of Science and Technology, Shenzhen 518055, China\\
              \email{rt.ryoji.tanabe@gmail.com, hisao@sustc.edu.cn (Corresponding author: Hisao Ishibuchi)}
}

\date{Received: date / Accepted: date}

\maketitle

\begin{abstract}


\begin{abstract}

%
A decomposition-based multi-objective evolutionary algorithm with a differential evolution variation operator (MOEA/D-DE) shows high performance on challenging multi-objective problems (MOPs).
The DE mutation consists of three key components: a mutation strategy, an index selection method for parent individuals, and a bound-handling method.
However, the configuration of the DE mutation operator that should be used for MOEA/D-DE has not been thoroughly investigated in the literature.
This configuration choice confuses researchers and users of MOEA/D-DE. 
To address this issue, we present a review of the existing configurations of the DE mutation operator in MOEA/D-DE and systematically examine the influence of each component on the performance of MOEA/D-DE.
Our review reveals that the configuration of the DE mutation operator differs depending on the source code of MOEA/D-DE.
In our analysis, a total of 30 configurations (three index selection methods, two mutation strategies, and five bound handling methods) are investigated on 16 MOPs with up to five objectives.
Results show that each component significantly affects the performance of MOEA/D-DE.
We also present the most suitable configuration of the DE mutation operator, which maximizes the effectiveness of MOEA/D-DE.



\end{abstract}

\end{abstract}

\section{Introduction}
\label{sec:introduction}


A multi-objective evolutionary algorithm (MOEA) is an efficient approach for solving multi-objective optimization problems (MOPs) \cite{Deb01}.
Since MOEAs use a set of individuals for the search, it is expected that well distributed nondominated solutions can be found by a single run.
MOEAs have been successfully applied to real-world problems, such as aerodynamic wing design problems \cite{OngNK03}, vehicle design problems \cite{LiaoLYZL08}, oil well problems \cite{LeOMJS13}, and groundwater monitoring design problems \cite{KollatRM13}.

A variation operator is used to generate a child in MOEAs.
Although genetic algorithm (GA) operators are frequently used in the evolutionary multi-objective optimization community, some recent studies report the superiority of differential evolution (DE) operators \cite{StornP97}. 
For example, the performance of NSGA-II \cite{DebAPM02}, SPEA2 \cite{ZitzlerLT01}, and IBEA \cite{ZitzlerK04} with GA and DE operators is investigated in \cite{TusarF07}.
The results reported in \cite{TusarF07} show that better nondominated solutions can be found by using the DE operator.
A further examination of the impact of  both of these operators on MOEAs is presented in  \cite{BezerraLS15}.
In \cite{YuanXW15}, the performance of NSGA-III \cite{DebJ14} with GA and DE operators is investigated on MOPs with more than four objectives.
A comparison of SMS-EMOA \cite{BeumeNE07} with various operators is performed in \cite{AugerBHTTW16b}.
The results presented in \cite{AugerBHTTW16b} show that the DE operator is more suitable than the GA operator for SMS-EMOA in most cases.

MOEA/D-DE \cite{LiZ09} is one of the most successful examples of the DE variation operator.
MOEA/D \cite{ZhangL07,TrivediSSG17} is an MOEA that decomposes a given MOP into multiple single-objective sub-problems and solves them simultaneously.
MOEA/D-DE is a variant of MOEA/D in which the GA operator is replaced by the DE operator\footnote{Strictly speaking, the differences between MOEA/D-DE \cite{LiZ09} and the original MOEA/D \cite{ZhangL07} are as follows: (i) the parent individuals are selected from the whole population with some probability, (ii) the number of individuals replaced by a child is restricted, and (iii) the DE variation operator is used in \cite{LiZ09}.}.
The results reported in \cite{LiZ09} show that MOEA/D-DE is capable of handling complicated Pareto solution sets.
Some improved MOEA/D-DE algorithms have been proposed in the literature, and representative examples include MOEA/D-DRA \cite{ZhangLL09}, MOEA/D-MAB \cite{LiFKZ14}, MOEA/D-STM \cite{LiZKLW14}, MOEA/D-GRA \cite{WangZZGJ16}, and MOEA/D-TPN \cite{JiangY16}.

In general, the term ``DE operator'' denotes a complex differential mutation operator and a crossover method \cite{StornP97}.
In the procedure of the DE operator, a mutant vector is first generated by applying the DE mutation operator to some individuals in the population.
Then, a child is generated by recombining the mutant vector and a parent individual.
Binomial crossover is widely used in MOEA/D-DE-type algorithms.
However, to handle nonseparability, the procedure of the crossover operator is not actually performed in most MOEA/D-DE-type algorithms by setting the crossover rate $\CR \in [0,1]$ to $1$ \cite{LiZ09}.

In contrast to the crossover method, there is large flexibility in implementing the DE mutation operator.
The DE mutation operator consists of the following three key components:
\begin{itemize}
\item A type of mutation strategy (e.g., the rand/1, rand/2, and current-to-rand/1 strategies),
\item An index selection method for parent individuals (i.e., how to select indices $r_1$, $r_2$, and $r_3$), and 
\item A bound-handling method to repair an infeasible solution. 
\end{itemize}

On the one hand, previous studies \cite{ArabasSW10,MontesRC06,QinHS09,WangLLLW16} reveal that the performance of DE algorithms for single-objective optimization is significantly influenced by the above-listed three components.
For example, the results presented in \cite{ArabasSW10} show that the choice of bound-handling methods has a large impact on the performance of the basic DE \cite{StornP97}.
For this reason, most researchers in the DE community carefully select each component of the DE mutation operator to maximize the performance of DE.




On the other hand, in contrast to the DE community, the details of the DE mutation operator have not received considerable attention in the evolutionary multi-objective optimization community.
Thus, the influence of configurations of the DE mutation operator on MOEA/D-DE has not been thoroughly investigated.
Since the configuration of the DE mutation operator that should be used is unclear, algorithm designers and users of MOEA/D-DE face confusion regarding the configuration choice.
In fact, different implementations are used in the uploaded source code of MOEA/D-DE.
For example, while the current/1 strategy is incorporated into MOEA/D-DE in the jMetal source code\footnote{\url{http://jmetal.sourceforge.net/}}, the rand/1 strategy is used in the MOEA Framework source code\footnote{\url{http://moeaframework.org/index.html}}.
Different comparison results could be obtained simply depending on the choice of a library if MOEA/D-DE in an MOEA library performs significantly better than that in another MOEA library.
This situation is undesirable for researchers.
Although users of MOEAs generally want to apply a well-performing algorithm to their MOPs, the best configuration of the DE mutation operator for MOEA/D-DE is unknown.


To address these issues, we present a review of the existing configurations of the DE mutation in MOEA/D-DE.
We also analyze the influence of each element on the effectiveness of MOEA/D-DE.
%
%
%
The purpose of this review is to clarify the current situation.
We examine the three components that can be found in the literature, source code, and MOEA libraries.
Then, we present a large-scale experimental study to investigate the effect of the three components on MOEA/D-DE.
A total of 30 configurations (three index selection methods, two mutation strategies, and five bound-handling methods) are investigated in our study.
The influence of each component on the performance of MOEA/D-DE is carefully examined in a component-wise manner.
We also suggest the best configuration of the DE mutation operator, which maximizes the performance of MOEA/D-DE.

The reminder of this paper is organized as follows:
Section \ref{sec:preliminaries} describes the background of MOPs and MOEA/D-DE.
Section \ref{sec:review} reviews existing configurations of the DE mutation operator in MOEA/D-DE.
Section \ref{sec:experimental_settings} describes experimental settings.
Section \ref{sec:experimental_results} shows results of MOEA /D-DE with various configurations of the DE mutation operator.
Finally, Section \ref{sec:conclusion} concludes this paper.

\section{Preliminaries}
\label{sec:preliminaries}

\subsection{Definition of MOPs}

A bound-constrained MOP, which is addressed in this paper, can be formulated as follows: 
\begin{align}
\label{eqn:mops}
\text{minimize  } \:\: &\vector{f}(\vector{x}) = \bigl(f_1 (\vector{x}), ..., f_M(\vector{x}) \bigr)^{\rm T},
\end{align}
where
$\mathbb{S}$ is the solution space.
$\vector{f}: \mathbb{S} \rightarrow \mathbb{R}^M$ is an objective function vector that consists of $M$ potentially conflicting objective functions, and $\mathbb{R}^M$ is the objective space.
In \eqref{eqn:mops}, $\vector{x} = (x_1, ..., x_D)^{\rm T}$ is a solution vector, and $\mathbb{S} = \prod^D_{j=1} [x^{\rm min}_j, x^{\rm max}_j]$ is the bound-constrained solution space where $x^{\rm min}_j \leq x_j \leq x^{\rm max}_j$ for each index $j \in \{1, ..., D\}$.


We say that $\vector{x}^1$ dominates $\vector{x}^2$  if and only if $f_i (\vector{x}^1) \leq f_i (\vector{x}^2)$ for all $i \in \{1, ..., M\}$ and $f_i (\vector{x}^1) < f_i (\vector{x}^2)$ for at least one index $i$.
Here, $\vector{x}^*$ is a Pareto-optimal solution if no $\vector{x} \in \mathbb{S}$ exists such that $\vector{x}$ dominates $ \vector{x}^*$.
 In this case, $\vector{f} (\vector{x}^*)$ is a Pareto-optimal objective vector.
The set of all $\vector{x}^*$ in $\mathbb{S}$ is the Pareto-optimal solution set, and the set of all $\vector{f}(\vector{x}^*)$ in $\mathbb{R}^M$ is the Pareto front.
In general, no solution can simultaneously minimize all objective functions $f_1, ..., f_M$ in an MOP.
Thus, the goal of multi-objective optimization is to find a set of nondominated solutions that are well distributed and close to the Pareto front in the objective space.


\subsection{MOEA/D-DE}
\label{sec:moeadde}

MOEA/D-type algorithms, including MOEA/D-DE \cite{LiZ09}, decompose an $M$-objective MOP defined in  \eqref{eqn:mops} into $\mu$ single-objective sub-problems $g_1(\vector{x} | \vector{w}^1), ..., g_{\mu}(\vector{x} | \vector{w}^{\mu})$ using a set of uniformly distributed weight vectors $\vector{W} = \{\vector{w}^1, ..., \vector{w}^{\mu}\}$ and a scalarizing function $\vector{g}: \mathbb{R}^M \rightarrow \mathbb{R}$, where $\mu$ is the population size, $\vector{w}^i = (w^i_1, ..., w^i_M)^{\rm T}$ for each $i \in \{1, ..., \mu\}$, and $\sum^M_{j=1} w^i_j = 1$.
%
For each $i \in \{1, ..., \mu\}$, an individual $\vector{x}^i$ is assigned to the $i$-th sub-problem.
MOEA/D-type algorithms attempt to find the optimal solutions of all sub-problems simultaneously.

 
Algorithm \ref{alg:moead09} shows the procedure of MOEA/D-DE.
%
%
After initialization (lines 1--3), the following steps are iteratively performed.
For each individual, an index list is selected for use in the mating and replacement selections (lines 5--9).
Then, the mating and reproduction operations are performed (lines 10--15).
Finally, the replacement selection is applied to the child and the individuals in the population (lines 17--20).
We explain each step of MOEA/D-DE in detail below.



%
%
At the beginning of the search, all individuals in the population $\vector{P} =\{ \vector{x}^{1}, ..., \vector{x}^{\mu}\}$ are randomly generated in the solution space (line 1).
For each sub-problem index $i \in \{1, ..., \mu\}$, an index list $\vector{B}^{i} = \{i_1, ..., i_{T}\}$ is initialized (lines 2-3): $\vector{B}^{i}$ consists of indices of the $T$ closest weight vectors to $\vector{w}^i$ in the weight vector space, where $T$ is the neighborhood size.
After initialization, the following steps (lines 5--20) are repeatedly applied to each sub-problem until a termination condition is satisfied.


\def\HiLi{\leavevmode\rlap{\hbox to \hsize{\color{black!15}\leaders\hrule height .8\baselineskip depth .5ex\hfill}}}

\IncMargin{0.5em}
\begin{algorithm}[t]
\scriptsize
\SetSideCommentRight
$t \leftarrow 1$, initialize the population $\vector{P} =\{ \vector{x}^{1}, ..., \vector{x}^{\mu}\}$\;
\For{$i \in \{1, ..., \mu\}$}{
 Set the neighborhood index list $\vector{B}^{i} = \{i_1, ..., i_{T}\}$\;
}
\While{$\textsf{\upshape{The termination criteria are not met}}$}{
  \For{$i \in \{1, ..., \mu\}$}{
    \uIf{${\rm rand} [0, 1] \leq \delta$} {
      $\vector{R} \leftarrow \vector{B}^{i}$\;
    }
    \Else{
      $\vector{R} \leftarrow \{1, ..., \mu\}$\;
    }
\HiLi Select parent indices from $\vector{R}$ with an index \HiLi  selection method (Subsection \ref{sec:parent_selections})\;
\HiLi Generate the mutant vector $\vector{v}^{i}$ using a mutation \HiLi strategy (Subsection \ref{sec:de_mutation_operator})\;
    \If{$\vector{v}^{i} \notin \mathbb{S}$} {
\HiLi Repair $\vector{v}^{i}$ using a bound-handling method \HiLi (Subsection \ref{sec:bchs})\;
    }
    Generate the child $\vector{u}^{i}$ by crossing $\vector{x}^{i}$ and $\vector{v}^{i}$\;
    Apply a GA mutation operator to $\vector{u}^i$\;
    $c \leftarrow 1$\;
    \While{$c \leq n^{\rm rep}$ $\textsf{\upshape{and}}$ $\vector{R}  \neq \emptyset$}{
      Randomly select an index $j$ from $\vector{R}$, and $\vector{R}  \leftarrow \vector{R} \backslash \{j\}$\;
      \If{$g(\vector{u}^i | \vector{w}^j, \vector{z}^*) \leq g(\vector{x}^j | \vector{w}^j, \vector{z}^*)$} {
        $\vector{x}^{j} \leftarrow \vector{u}^i$, $c \leftarrow c + 1$\;       
      }
    }
  }
 $t \leftarrow t + 1$\;
}
\caption{The procedure of MOEA/D-DE}
\label{alg:moead09}
\end{algorithm}\DecMargin{0.5em}

For each $i$, a set of individual indices $\vector{R}$ is set to $\vector{B}^i$  with a probability of $\delta \in [0,1]$ or $\{1, ..., \mu\}$ with a probability of $1-\delta$ (lines 6--9).
The function ${\rm rand}[0,1]$ in line 6 is a randomly chosen value in the range $[0,1]$.
%
After $\vector{R}$ has been determined, MOEA/D-DE generates a mutant vector $\vector{v}^i$ (lines 10--13).
First, individual indices $\{r_1, r_2, ...\}$ are randomly selected from $\vector{R}$ using an index selection method (line 10).
Next, $\vector{v}^i$ is generated by applying a mutation strategy to the selected individuals $\{\vector{x}^{r_1}, \vector{x}^{r_2}, ...\}$  (line 11).
If an element of $\vector{v}^i$ violates a corresponding bound constraint (i.e., $\vector{v}^{i} \notin \mathbb{S}$), then a bound-handling method is applied to it such that  $\vector{v}^{i} \in \mathbb{S}$ (line 13).
The details of each procedure (the index selection, the mutation strategy, and the bound-handling method) are described in Subsections  \ref{sec:parent_selections}, \ref{sec:de_mutation_operator}, and \ref{sec:bchs}, respectively.


After $\vector{v}^{i}$ is repaired by a bound-handling method (if needed), a child $\vector{u}^{i}$ is generated by recombining $\vector{x}^{i}$ and $\vector{v}^{i}$ (line 14).
Binomial crossover \cite{StornP97}, which is the most basic crossover method in DE, is defined as follows:
\begin{align}
  \label{eqn:bin}
  u^i_j = \begin{cases}
    v^i_j & \:  {\rm if} \: {\rm rand}[0,1] \leq \CR \: {\rm or} \: j = j^{\rm rand}\\
    x^i_j & \:  {\rm otherwise}
  \end{cases},
\end{align}
where the crossover rate $\CR \in [0,1]$ in  \eqref{eqn:bin} controls the number of inherited variables from $\vector{x}^i$ to $\vector{u}^i$.
The decision variable index $j^{\rm rand}$ in \eqref{eqn:bin} is randomly selected from $\{1, ...,  D\}$.
Since binomial crossover only exchanges elements between the parent individual $\vector{x}^i$ and the mutant vector $\vector{v}^i$, the child $\vector{u}^i$ always satisfies the bound constraints as long as $\vector{x}^i$, $\vector{v}^i \in \mathbb{S}$.

In MOEA/D-DE, a GA mutation operator is applied to the child $\vector{u}^i$ (line 15).
The following polynomial mutation, which is commonly used in the evolutionary multi-objective optimization community, is performed for each $j \in \{1, ..., D\}$:
\begin{align}
  \label{eqn:pm}
  u^i_j = \begin{cases}
    u^i_j + \sigma_j \, (x^{\rm max}_{j} - x^{\rm min}_{j}) & \:  {\rm if} \: {\rm rand}[0,1] \leq p^{\rm mut}\\
    u^i_j & \:  {\rm otherwise}
  \end{cases},
\end{align}
where $p^{\rm mut} \in [0,1]$ is the mutation rate.
If $u^i_{j} \notin [x^{\rm min}_j, x^{\rm max}_j]$, then it is replaced by the closest value ($x^{\rm min}_j$ or $x^{\rm max}_j$).
The amount of perturbation $\sigma_j$ in  \eqref{eqn:pm} is defined as follows:
%
\begin{align}
  \label{eqn:pm_sigma}
  \sigma_{j} = \begin{cases}
    (2 \, {\rm rand}[0,1])^{1/(\eta^{\rm mut} +1)} - 1 & {\rm if} \: {\rm rand}[0,1] \leq 0.5\\
    1 - (2 - 2 \, {\rm rand}[0,1])^{1/(\eta^{\rm mut} +1)} &  {\rm otherwise}
  \end{cases},
\end{align}
%
where $\eta^{\rm mut} > 0$ in \eqref{eqn:pm_sigma} is the distribution index.

After $\vector{u}^i$ is generated, the replacement procedure is performed using a predefined scalarizing function $g$ (lines 17--20).
First, an index $j$ is randomly selected from $\vector{R}$, and $j$ is removed from $\vector{R}$ (line 18).
Then, the individual $\vector{x}^j$ is compared with the child $\vector{u}^i$ based on $g$ and the weight vector $\vector{w}^j$ (line 19).
If $\vector{u}^i$ is better than $\vector{x}^j$ according to their scalarizing function values, then $\vector{x}^j$ is replaced with $\vector{u}^i$ (line 20).
Unlike the original MOEA/D proposed in 2007 \cite{ZhangL07}, the number of individuals replaced by the child is limited to $n^{\rm rep} > 0$ in MOEA/D-DE (line 17). 
$c$ is used to count the number of individuals replaced by $\vector{u}^i$.
When $c$ reaches the maximum number of replacements $n^{\rm rep}$, the replacement procedure terminates (line 17).




Since the replacement criterion is based on the scalarizing function $g$ (line 19), the performance of MOEA/D-type algorithms significantly depends on $g$ \cite{IshibuchiAN15}.
Although there are a number of scalarizing functions, as reviewed in \cite{Pescador-RojasG17}, the following Tchebycheff function ($g^{\rm tch}$) \cite{ZhangL07} is used in MOEA/D-DE:
\begin{align}
\label{eqn:Chebyshev-mul}
g^{\rm tch}(\vector{x} | \vector{w}, \vector{z}^*) &= \max_{i \in \{1, ..., M\}} \{ w_i |f_i (\vector{x}) - z^*_i|  \},
\end{align}
where $g^{\rm tch}$ in  \eqref{eqn:Chebyshev-mul} should be minimized.
The $\vector{z}^* = (z^*_1, ..., $ $z^*_M)^{\rm T}$ is the ideal point.
Since obtaining the true ideal point $\vector{z}^*$ for a given MOP is difficult, its approximated point, which consists of the minimum function value for each objective $f_i$ ($i \in \{1, ..., M\}$) found during the search process, is typically used. 

  \begin{table*}[t]
\centering
  \caption{Configurations of the DE mutation operator in the source code of MOEA/D-DE and its variants. Eight configuration numbers from \#A to \#H are used in Subsection \ref{sec:results_existing_codes}.}
  \label{tab:moead_de_source_code}
\begin{threeparttable}
\begin{tabular}{llllllc}
\midrule
\raisebox{1em}{MOEAs} & \raisebox{1em}{Languages} & \raisebox{1em}{Web sites} & \raisebox{0.5em}{\shortstack{Mutation\\strategies}} & \shortstack[l]{Index\\selection\\methods} & \shortstack[l]{Bound-\\handling\\methods} & \raisebox{0.5em}{\shortstack{Configuration\\number \#}}\\
\toprule
\shortstack[l]{MOEA/D-DE\\described in \cite{LiZ09}} &  & & \raisebox{0.5em}{current/1} &  \raisebox{0.5em}{WR} & \raisebox{0.5em}{reinitialization} & \#A\\\midrule
MOEA/D-DE & C++ & MOEA/D homepage\tnote{a} & current/1 &  WPR & replacement & \#B\\\midrule
MOEA/D-DE & MATLAB & MOEA/D homepage\tnote{a} & rand/1 &  WOR & replacement & \#C\\\midrule
MOEA/D-DE & Java & MOEA/D homepage\tnote{a} & rand/1 &  WOR & replacement & \#C\\\midrule
\midrule
\shortstack[l]{MOEA/D-DE\\MOEA/D-DRA} & \raisebox{0.5em}{Java} & \raisebox{0.5em}{jMetal 4.5\tnote{b}} & \raisebox{0.5em}{current/1} &  \raisebox{0.5em}{WPR} & \raisebox{0.5em}{replacement} & \raisebox{0.5em}{\#B}\\\midrule
MOEA/D-DRA & Java & MOEA Framework\tnote{c} & rand/1 &  WPR & replacement & \#D\\\midrule
MOEA/D-DE & C++ &  PaGMO\tnote{d} & current/1 &  WOR & r-reflection & \#E\\\midrule
MOEA/D-DE & MATLAB & PlatEMO\tnote{e} & current/1 &  WOR & no method & -\\\midrule
\midrule
MOEA/D-STM & Java & Li's website\tnote{f} & current/1 &  WPR & replacement & \#B\\\midrule
MOEA/D-STM & MATLAB & Li's website\tnote{f} & current/1 &  WOR & replacement & \#F\\\midrule
MOEA/D-DE & C & COCO website\tnote{g} & rand/1 &  WOR & r-reflection & \#G\\\midrule
MOEA/D-DRA & C++ & MOEA/D homepage\tnote{a} & current/1 &  WPR & r-reflection & \#H\\\midrule
ENS-MOEA/D & MATLAB & Suganthan's website\tnote{h}& current/1 &  WPR & r-reflection & \#H\\\midrule
MOEA/D-TPN & C & Yang's website\tnote{i} & current/1 & WPR & r-reflection & \#H\\\midrule
\end{tabular}
\begin{tablenotes}
\item[a] \url{http://dces.essex.ac.uk/staff/zhang/webofmoead.htm}
\item[b] \url{http://jmetal.sourceforge.net/}
\item[c] \url{http://moeaframework.org/index.html}
\item[d] \url{https://esa.github.io/pagmo2/index.html}
\item[e] \url{http://bimk.ahu.edu.cn/index.php?s=/Index/Software/index.html}
\item[f] \url{http://www.cs.bham.ac.uk/~likw/publications.html}
\item[g] \url{http://coco.gforge.inria.fr/doku.php?id=mo-gecco2015}
\item[h] \url{http://www3.ntu.edu.sg/home/EPNSugan/index_files/}
\item[i]  \url{http://www.tech.dmu.ac.uk/~syang/publications.html}
\end{tablenotes}
\end{threeparttable}
 \end{table*}

\section{A review of existing configurations of the DE mutation operator in MOEA/D-DE}
\label{sec:review}


Here, we review the existing configurations of the DE mutation operator in MOEA/D-DE.
We also explain why many configurations exist in the source code of MOEA/D-DE.
%
Table \ref{tab:moead_de_source_code} shows 15 configurations of the DE mutation operator in the source code of MOEA/D-DE and its variants.
We downloaded the source code from each website in Table \ref{tab:moead_de_source_code} and carefully checked how the differential mutation was implemented.
The components of the DE mutation described here include two mutation strategies, three index selection methods for parent individuals, and five bound-handling methods. 
These components are explained in Subsections \ref{sec:de_mutation_operator}, \ref{sec:parent_selections}, and \ref{sec:bchs}, respectively.



\subsection{Two DE mutation strategies in MOEA/D-DE}
\label{sec:de_mutation_operator}

For each target individual $\vector{x}^i$ ($i \in \{1, ..., \mu\}$) in the population $\vector{P}$, MOEA/D-DE performs the differential mutation to generate a mutant vector $\vector{v}^i$ (line 11 in Algorithm \ref{alg:moead09}).
A number of mutation strategies exist in DE  \cite{DasMS16,DasS11}.
Among these strategies, the following rand/1 strategy is the most basic mutation strategy:
\begin{align}
  \label{eqn:rand_1}
  \vector{v}^{i} = \vector{x}^{r_1} + F \: (\vector{x}^{r_2} - \vector{x}^{r_3}),
\end{align}
where $r_1$, $r_2$, and $r_3$ are randomly selected from a set of individual indices $\vector{R}$ (see lines 6--9 in Algorithm \ref{alg:moead09}) according to the index selection method, which is explained in Subsection \ref{sec:parent_selections} later.
The rand/1 mutation strategy is widely used in DE algorithms for single-objective optimization and multi-objective optimization.
For example, the rand/1 operator is incorporated into representative multi-objective DE algorithms, such as GDE3 \cite{KukkonenL05} and DEMO \cite{RobicF05}.


Although the rand/1 operator is commonly used in single- and multi-objective DE, the current/1 mutation strategy is incorporated into MOEA/D-DE.
Note that the term ``current/1'' is introduced in \cite{GongWCY17} and is not used in the original MOEA/D-DE paper \cite{LiZ09}.
The current/1 mutation strategy is defined as follows:
\begin{align}
  \label{eqn:current_1}
  \vector{v}^{i} = \vector{x}^{i} + F \: (\vector{x}^{r_1} - \vector{x}^{r_2}),
\end{align}
where $r_1$ and $r_2$ are randomly selected from $\vector{R}$.
Although the base vector is a randomly selected individual from $\vector{R}$ in the rand/1 strategy, it is always identical to the target individual $\vector{x}^{i}$ in the current/1 strategy.
Compared to the rand/1 strategy, the mutant vector $\vector{v}^{i}$ generated by the current/1 strategy is generally close to $\vector{x}^{i}$ \cite{GongWCY17}.



As mentioned above, MOEA/D-DE uses the current/1 strategy for the mutant vector generation.
However, in the MOEA/D-DE paper \cite{LiZ09}, it is not explicitly stated that the current/1 mutation strategy in \eqref{eqn:current_1} is incorporated into MOEA/D-DE.
Instead, in \cite{LiZ09}, it is reported that the rand/1 strategy is used in MOEA/D-DE (equation (6) on page 9 in \cite{LiZ09}), but the randomly chosen index $r_1$ is replaced by the target individual index $i$ (Step 2.2 on page 9 in \cite{LiZ09}).
The modified rand/1 strategy is identical to the current/1 strategy.


The complicated description in \cite{LiZ09} is somewhat difficult to understand.
At first glance, one may think that the well-known rand/1 strategy is used in MOEA/D-DE.
Unless readers have carefully read all the descriptions of the DE variation operator in MOEA/D-DE, then they will not notice that the current/1 strategy  is actually used in MOEA/D-DE.
In fact, the descriptions in \cite{LiZ09} have confused some researchers.
Consequently, the rand/1 strategy is incorporated into MOEA/D-DE in some source code.
For example, as shown in Table \ref{tab:moead_de_source_code}, MOEA/D-DE in MOEA Framework and the source code of \cite{BrockhoffTH15} use the rand/1 strategy, rather than the current/1 strategy.
Additionally, some authors describe that the rand/1 mutation strategy was used in MOEA/D-DE-type algorithms in their articles, but the current/1 strategy was actually used in their source code.
MOEA/D-TPN \cite{JiangY16} in Table \ref{tab:moead_de_source_code} is such an example.
Since the rand/1 and current/1 strategies are essentially different, it is expected that the performance of MOEA/D-DE with the two strategies is  different.

\begin{figure*}[t]
\centering
\begin{lstlisting}[language=C++,basicstyle=\ttfamily\scriptsize, numbers=left,frame=single]
void CMOEAD::matingselection(vector<int> &list, int cid, int size, int type){
  // list : the set of the indexes of selected mating parents
  // cid  : the id of current subproblem
  // size : the number of selected mating parents
  // type : 1 - neighborhood; otherwise - whole population
  int ss   = population[cid].table.size(), r, p;
  while(list.size()<size)
    {
      if(type==1){
	r = int(ss*rnd_uni(&rnd_uni_init));
	p = population[cid].table[r];
      }
      else
	p = int(population.size()*rnd_uni(&rnd_uni_init));

      bool flag = true;
      for(int i=0; i<list.size(); i++)
	{
	  if(list[i]==p) // p is in the list
	    { 
	      flag = false;
	      break;
	    }
	}

      if(flag) list.push_back(p);
    }
}
 \end{lstlisting}
\caption{
The ``matingselection'' function in the original C++ code of MOEA/D-DE, which is available on the authors' website.
To implement the WOR selection method, line 19 should have been ``if(list[i]==p \&\& cid==p)''.
}
\label{fig:bug_in_matingselection}
\end{figure*}

\subsection{Index selection methods for the parent individuals}
\label{sec:parent_selections}

Here, we explain three index selection methods (WOR, WR, and WPR methods) for the parent individuals in the DE mutation operator.
Algorithms \ref{alg:wo-replacement}, \ref{alg:w-replacement}, and \ref{alg:wp-replacement} show the procedures of the WOR, WR, and WPR methods for selecting two individual indices $r_1$ and $r_2$ for the current/1 mutation strategy.
In Algorithm \ref{alg:wo-replacement} (WOR), $r_1$ and $r_2$ are randomly selected such that $r_1 \neq r_2$, $r_1 \neq i$, and $r_2 \neq i$.
In contrast, $i$, $r_1$, and $r_2$ can be the same index (i.e., $r_1 = r_2 = i$) in Algorithm \ref{alg:w-replacement} (WR).
In Algorithm \ref{alg:wp-replacement} (WPR), $r_1$ and $r_2$ differ from each other, but either of them can be equal to $i$.
In the following, the three index selection methods are described in detail.


In the original DE paper \cite{StornP97}, the parent individual indices $r_1$, $r_2$, and $r_3$ for the rand/1 strategy are randomly selected from $\{1, ..., \mu\} \backslash \{i\}$ such that they differ from each other (i.e., $r_1 \neq r_2$, $r_1 \neq r_3$, $r_2 \neq r_3$, and $r_j \neq i$ for $j \in \{1, 2, 3\}$).
That is, individual indices are selected without replacement.
In this paper, the traditional index selection method in DE is called a WOR method (Algorithm \ref{alg:wo-replacement}).
Although the WOR selection method is generally used in DE, such a restriction is not described in the original MOEA/D-DE paper \cite{LiZ09}.
In this case, the selected indices are possibly equal to each other and also equal to $i$.
The method that selects individual indices with replacement is called a WR method (Algorithm \ref{alg:w-replacement}).


\IncMargin{0.5em}
\begin{algorithm}[t]
\footnotesize
\SetKwInOut{Input}{input}\SetKwInOut{Output}{output}
\SetSideCommentRight
 \Do{$r_1 = i$}{
   Randomly select $r_1$ from $\vector{R}$\;
}
 \Do{$r_2 = i$ $\textsf{\upshape{and}}$ $r_2 = r_1$}{
   Randomly select $r_2$ from $\vector{R}$\;
}
\caption{The WOR selection method.}
\label{alg:wo-replacement}
\end{algorithm}\DecMargin{0.5em}


\IncMargin{0.5em}
\begin{algorithm}[t]
\footnotesize
\SetKwInOut{Input}{input}\SetKwInOut{Output}{output}
\SetSideCommentRight
Randomly select $r_1$ from $\vector{R}$\;
Randomly select $r_2$ from $\vector{R}$\;
\caption{The WR selection method.} 
\label{alg:w-replacement}
\end{algorithm}\DecMargin{0.5em}

\IncMargin{0.5em}
\begin{algorithm}[t]
\footnotesize
\SetKwInOut{Input}{input}\SetKwInOut{Output}{output}
\SetSideCommentRight
Randomly select $r_1$ from $\vector{R}$\;
\Do{$r_2 = r_1$}{
  Randomly select $r_2$ from $\vector{R}$\;
}
\caption{The WPR selection method.}
\label{alg:wp-replacement}
\end{algorithm}\DecMargin{0.5em}

However, in the original C++ code of MOEA/D-DE, $r_1$ and $r_2$ for the current/1 strategy are selected from a set of individual indices $\vector{R}$ (see Subsection \ref{sec:moeadde}) such that they differ from each other, but there is no restriction about $i$.
Figure \ref{fig:bug_in_matingselection} shows the \verb|matingselection| function in the original C++ source code of MOEA/D-DE, which is available on the authors' website.
The \verb|matingselection| function is for selecting indices for the differential mutation.
We do not explain the procedure of the \verb|matingselection| function in detail, but there is a bug in line 19 in Figure \ref{fig:bug_in_matingselection}.
The one-dimensional vector \verb|list| is for storing indices $\{r_1, r_2, ...\}$ that have been selected in the \verb|matingselection| function.
In line 19, \verb|p| is a randomly selected candidate that is able to enter \verb|list|. 
Since it is checked whether \verb|p| exists in  \verb|list|, \verb|p| may equal  \verb|cid|, where \verb|cid| is the index of the target sub-problem and identical to $i$ in Algorithm \ref{alg:moead09}.
Thus, while it is ensured that $r_1$ differs from $r_2$, either of the two indices may equal $i$.
The selection method with partial replacement used in the original MOEA/D-DE code is called a WPR method (Algorithm \ref{alg:wp-replacement}) in this paper.
The WPR selection method has never been found in the DE literature.


We carefully checked the index selection method in each source code in Table \ref{tab:moead_de_source_code}.
As shown in Table \ref{tab:moead_de_source_code}, it appears that the WR method has hardly been used in MOEA/D-DE.
The WPR selection method is used for MOEA/D-DE in some MOEA packages (e.g., jMetal and MOEA Framework), which may be because they were based on the original MOEA/D-DE source code.
The WOR method is used in other source code (e.g., PaGMO and PlatEMO), which is because the WOR is the most standard index selection method in the DE community.
If researchers attempt to implement MOEA/D-DE without referring to its original C++ source code, most of them are likely to incorporate the WOR method into MOEA/D-DE for this reason.


Wang et al. \cite{WangLLLW16} investigate the impact of the index selection method on the performance of DE algorithms for single-objective optimization.
Their results show that the performance of some DE algorithms can be significantly improved or degraded by replacing the traditional WOR method with the WR method.
Since removing the restriction $r_1 \neq r_2$ increases the probability that a child is generated near the base vector, the use of the WR method makes the search more greedy \cite{WangLLLW16}.
Thus, it is expected that the performance of MOEA/D-DE is affected by the choice of index selection method.



\subsection{Bound constraint-handling methods}
\label{sec:bchs}

It is not always ensured that the mutant vector $\vector{v}$ generated by the differential mutation is in the solution space $\mathbb{S} = \Pi^D_{j=1} [x^{\rm min}_j, x^{\rm max}_j]$.
When an element of the mutant vector $v_j$ is out of the bound ($j \in \{1, ..., D\}$), a bound-handling method must be applied to $\vector{v}$ such that $\vector{v} \in \mathbb{S}$ (line 13 in Algorithm \ref{alg:moead09}).
In our study, we consider the following five bound-handling methods (resampling, replacement, reinitialization, reflection, and r-reflection methods), which are commonly used in the DE literature:

\noindent {\bf $\bullet$ Resampling}:
The mutant vector $\vector{v}$ is repeatedly generated using the differential mutation with different parent individuals until all elements of $\vector{v}$ satisfy the bound constraints.

\noindent {\bf $\bullet$ Replacement}:
In the case where $v_j$ is out of the bound, it is replaced with the corresponding minimum/maximum value ($x^{\rm min}_j$ or $x^{\rm max}_j$) as \eqref{eqn:replacement}:
\begin{align}  
  \label{eqn:replacement}
  v_j = \begin{cases}
    x^{\rm min}_j &   {\rm if} \: v_j  < x^{\rm min}_j\\
    x^{\rm max}_j &   {\rm if} \: v_j  > x^{\rm max}_j\\
    v_j &   {\rm otherwise}\\    
  \end{cases}.
\end{align}
%


\noindent {\bf $\bullet$ Reinitialization}:
When $v_j \notin [x^{\rm min}_j, x^{\rm max}_j]$, it is reinitialized in the range $[x^{\rm min}_j, x^{\rm max}_j]$ as \eqref{eqn:reinitialization}:
\begin{align}  
  \label{eqn:reinitialization}
  v_j = \begin{cases}
    (x^{\rm max}_j - x^{\rm min}_j) \, {\rm rand}[0,1] + x^{\rm min}_j &   \parbox[t]{0.3\textwidth}{if $v_j  < x^{\rm min}_j$\\ or $v_j  > x^{\rm max}_j$}\\
    v_j &   {\rm otherwise}
  \end{cases}\hspace{-9.3em}.
\end{align}

\noindent {\bf $\bullet$ Reflection}:
If $v_j$ violates the $j$-th bound constraint, it is reflected as much as it exceeded the minimum/maximum value as \eqref{eqn:reflection}:
\begin{align}  
  \label{eqn:reflection}
  v_j = \begin{cases}
    x^{\rm min}_j  +  (x^{\rm min}_j - v_j) &   {\rm if} \: v_j  < x^{\rm min}_j\\
    x^{\rm max}_j +  (x^{\rm max}_j  - v_j) &   {\rm if} \: v_j  > x^{\rm max}_j\\
    v_j &   {\rm otherwise}\\    
  \end{cases}.
\end{align}

\noindent {\bf $\bullet$ Randomized reflection (r-reflection)}:
This method is an alternative version of the reflection method.
Unlike the reflection method, the amount of reflection is randomly determined as \eqref{eqn:reflection}:
\begin{align}  
  \label{eqn:reflection}
  v_j = \begin{cases}
    x^{\rm min}_j  + {\rm rand}[0,1] \, (x^{\rm min}_j - v_j) &   {\rm if} \: v_j  < x^{\rm min}_j\\
    x^{\rm max}_j + {\rm rand}[0,1] \, (x^{\rm max}_j  - v_j) &   {\rm if} \: v_j  > x^{\rm max}_j\\
    v_j &   {\rm otherwise}\\    
  \end{cases}.
\end{align}

Table \ref{tab:moead_de_source_code} shows that the replacement method is used in the original C++ source code and most source code for MOEA/D-DE.
In addition to the replacement method, the r-reflection method is used in some implementations, as shown in Table \ref{tab:moead_de_source_code}.
Table \ref{tab:moead_de_source_code} indicates that the reinitialization, reflection, and resampling methods have not been incorporated into MOEA/D-DE.
Although it is described in \cite{LiZ09} that the reinitialization method is used as the repair method (Step 2.3 on page 9 in \cite{LiZ09}), the replacement strategy is actually incorporated into the original C++ source code.
While the reflection method is one of the most popular repair methods in DE algorithms for single-objective optimization, it has not been used in MOEA/D-DE.
It was shown in \cite{ArabasSW10} that the resampling method is most suitable for DE for single-objective optimization.
However, the resampling method has not received much attention in the evolutionary multi-objective optimization community.


The use of a repair operator in evolutionary algorithms, including DE algorithms, influences their performance \cite{ArabasSW10,HansenNGK09,HelwigBM13,Wessing13}.
The impact of the bound-handling method on the basic DE algorithm is investigated in \cite{ArabasSW10}.
The results reported in \cite{ArabasSW10} indicate that the choice of repair method significantly influences the performance of the basic DE \cite{StornP97} for single-objective continuous optimization.
Thus, it is expected that the performance of MOEA/D-DE also depends on the choice of bound-handling method.

\section{Experimental settings}
\label{sec:experimental_settings}

%



This section describes our experimental settings.
Results are reported in Section \ref{sec:experimental_results}.

\subsection{Test problems}

The seven DTLZ \cite{DebTLZ05} and nine WFG  \cite{HubandHBW06} test problems with $M \in \{2, 3, 4, 5\}$ were used in our study.
Table \ref{tab:wfg_properties} summarizes their properties.
Note that the Pareto fronts of the DTLZ5, DTLZ6, and WFG3 problems are partially degenerate \cite{IshibuchiMN16}.
According to \cite{DebTLZ05}, for the DTLZ problems, the number of position variables $k$ was set to $k=5$ for the DTLZ1 problem, $k=20$ for the DTLZ7 problem, and $k=10$ for the other  DTLZ problems, where the number of variables is $D=M+k-1$.
Additionally, as suggested in \cite{HubandHBW06}, for the WFG test problems, the number of position variables  $k$ was set to $k = 2 \, (M-1)$, and the number of distance variables $l$ was set to $l=20$, where $D = k+l$.

\begin{table}[t]
\begin{center}
  \caption{Properties of the DTLZ and WFG test problems.
``Mult.'' and ``Nonsep.'' represent the multimodality and the nonseparability, respectively.
  }
{\scriptsize
  \label{tab:wfg_properties}
\begin{tabular}{llccc}
\midrule
Problem & Pareto front & Mult. & Nonsep. & Others\\
\toprule
DTLZ1 & Linear & $\checkmark$ & & \\\midrule
DTLZ2 & Nonconvex &  & & \\\midrule
DTLZ3 & Nonconvex & $\checkmark$ & & \\\midrule
DTLZ4 & Nonconvex & & & Biased\\\midrule
\raisebox{0.5em}{DTLZ5} & \shortstack[l]{Partially\\degenerate} & & & \\\midrule
\raisebox{0.5em}{DTLZ6} & \shortstack[l]{Partially\\degenerate} & & & \\\midrule
DTLZ7 & Disconnected & $\checkmark$ & & \\\midrule
\midrule
WFG1 & Mixed &  & & Biased\\\midrule
WFG2 & Disconnected & $\checkmark$ & $\checkmark$ & \\\midrule
\raisebox{0.5em}{WFG3} & \shortstack[l]{Partially\\degenerate} &  & \raisebox{0.5em}{$\checkmark$} & \\\midrule
WFG4 & Nonconvex & $\checkmark$ &  & \\\midrule
WFG5 & Nonconvex &  &  & Deceptive\\\midrule
WFG6 & Nonconvex &  & $\checkmark$ & \\\midrule
WFG7 & Nonconvex &  &  & Biased\\\midrule
WFG8 & Nonconvex &  & $\checkmark$ & Biased\\\midrule
\raisebox{0.5em}{WFG9} & \raisebox{0.5em}{Nonconvex} & \raisebox{0.5em}{$\checkmark$} & \raisebox{0.5em}{$\checkmark$} & \shortstack{Deceptive\\Biased}\\\midrule
\end{tabular}
}
\end{center}
\end{table}

\subsection{Performance indicators}
\label{sec:performance_indicators}

The hypervolume indicator \cite{ZitzlerTLFF03} was used for evaluating the quality of a set of obtained nondominated solutions $\vector{A}$.
Before calculating the hypervolume value, the objective vector $\vector{f} (\vector{x})$ of each $\vector{x} \in \vector{A}$ was normalized using the ideal point $\vector{z}^{\rm ideal} = (z^{\rm ideal}_1, ..., z^{\rm ideal}_M)^{\rm T}$ and the nadir point $\vector{z}^{\rm nadir} = (z^{\rm nadir}_1, ..., z^{\rm nadir}_M)^{\rm T}$ of each problem:
\begin{align}
f^{\rm normalized}_i(\vector{x}) = \frac{f_i(\vector{x}) - z^{\rm ideal}_i}{z^{\rm nadir}_i - z^{\rm ideal}_i}, 
\end{align}
where $f^{\rm normalized}_i(\vector{x})$ is the $i$-th normalized objective value ($i \in \{1, ..., M\}$).
The $i$-th element $z^{\rm ideal}_i$ of $\vector{z}^{\rm ideal}$ is the minimum value of the $i$-th objective function of the true Pareto front.
Conversely, the $i$-th element $z^{\rm nadir}_i$ of $\vector{z}^{\rm nadir}$ is the maximum value of the $i$-th objective function of the true Pareto front.
That is, the objective space of each test problem is normalized such that the ideal point and the nadir point in the normalized objective space are $(0, 0, ..., 0)^{\rm T}$ and $(1, 1, ..., 1)^{\rm T}$, respectively.
According to \cite{IshibuchiSMN16}, the reference point for the hypervolume calculation was set to $(1.1, ..., 1.1)^{\rm T}$.
In this setting, the hypervolume value is in the range $[0, 1.1^M]$.

The average performance score (APS) \cite{BaderZ11} was used to aggregate the results on various problems.
Suppose that $n$ algorithms $A_1, ..., A_n$ are compared for a given problem instance based on the hypervolume values obtained in multiple runs.
For each $i \in \{1, ..., n\}$ and $ j \in \{1, ..., n\} $ $\backslash \{i\}$, if $A_j$ significantly outperforms $A_i$ using the Wilcoxon rank-sum test with $p < 0.05$, then $\delta_{i,j} = 1$; otherwise, $\delta_{i,j} = 0$.
The performance score $P(A_i)$ is defined as follows: $P(A_i) = \sum^{n}_{ j \in \{1, ..., n\} \backslash \{i\}} \delta_{i,j}$.
The score $P(A_i)$ represents the number of algorithms that outperform $A_i$. 
The APS value of $A_i$ is the average of the $P(A_i)$ values for all the considered problem instances.
In other words, the APS value of $A_i$ represents how good (relatively) the performance of $A_i$ is among the $n$ algorithms on average over all problem instances. 
A small APS value indicates that the performance of the target algorithm is better than other compared algorithms.

\begin{table}[t]
\renewcommand{\arraystretch}{0.8}
\centering
  \caption{\small 
Example of the APS calculation.
%
}
{\footnotesize
  \label{tab:aps_example}
\scalebox{1}[1]{
\begin{tabular}{ccccccccc}
  \midrule
  Problem & $A_1$ & $A_2$ & $A_3$\\
\toprule
$I_1$ & 2 & 1 & 0\\
$I_2$ & 0 & 0 & 2\\
$I_3$ & 1 & 0 & 1\\
$I_4$ & 2 & 0 & 1\\
\midrule
APS & 1.25 (5/4) & 0.25 (1/4) & 1 (4/4)\\
\midrule
\end{tabular}
   \color{black}
}
}
\end{table}

Table \ref{tab:aps_example} shows a simple example of the APS calculation.
Let us consider the APS values of three algorithms $A_1$, $A_2$, and $A_3$ on four problems $I_1$, $I_2$, $I_3$, and $I_4$.
Each element of Table \ref{tab:aps_example} shows a performance score value of the corresponding algorithm.
The performance score value represents the number of algorithms that outperform the corresponding algorithm.
For example, $A_1$ is outperformed by two algorithms on $I_1$ and $I_4$.
$A_3$ is not outperformed by any other algorithms on $I_1$.
The APS value for each algorithm is calculated by dividing the total performance score value (e.g., $5$ for $A_1$) by the number of problems (i.e., $4$ for all algorithms).
In this example, the APS values of $A_1$, $A_2$, and $A_3$ are 1.25, 0.25, and 1, respectively.
It can be concluded that $A_2$ is the best performer according to the calculated APS values.

\subsection{Control parameters of MOEA/D-DE}
\label{sec:params_moeadde}


We used the source code of MOEA/D-DE downloaded from the jMetal website for our experiments.
All the control parameters (except for the population size $\mu$) for MOEA/D-DE were set according to the original study \cite{LiZ09}. 
For the DE operators, the scale factor $F$ and the crossover rate $\CR$ were set to $0.5$ and $1$, respectively.
For the polynomial mutation, $p^{\rm mut}$ and $\eta^{\rm mut}$ were set to $1/D$ and $20$, respectively. 
The other parameters of MOEA/D-DE were set as follows: $T=20$, $n^{\rm rep}=2$, and $\delta=0.9$.
The population size $\mu$ was set to $200$, $210$, $220$, and $210$ for $M = 2, 3, 4,$ and $5$, respectively.
The simple normalization strategy described in \cite{ZhangL07} was introduced into MOEA/D-DE to handle differently scaled objective function values.
The maximum number of function evaluations was set to $100\,000$ for all test problems, and $51$ independent runs were performed.


\newcommand{\chodoiikanjinosize}{\fontsize{7.4}{0}\selectfont}

\begin{table}[t]
\centering
  \caption{\small 
Overall performance of MOEA/D-DE with the three index selection methods on the 16 problems.
The APS value at the final iteration is shown in the tables (lower is better).
The numbers in parentheses indicate the ranks of the three configurations based on their APS values.
}
  \label{tab:results_psm}
\subfloat[current/1]{    
\begin{tabular}{ccccccccc}
\midrule
$M$ & WOR & WR & WPR\\
\toprule
$2$ & 0.88 (3)  & 0.12 (2)  & {\chodoiikanjinosize \textbf{0.06}} (1)  \\
$3$ & 0.56 (3)  & {\chodoiikanjinosize \textbf{0.06}} (1)  & 0.31 (2)  \\
$4$ & {\chodoiikanjinosize \textbf{0.25}} (1)  & {\chodoiikanjinosize \textbf{0.25}} (1)  & 0.31 (3)  \\
$5$ & 0.38 (3)  & {\chodoiikanjinosize \textbf{0.00}} (1)  & 0.12 (2)  \\
\midrule
\end{tabular}
}
\\
\subfloat[rand/1]{    
\begin{tabular}{ccccccccc}
\midrule
$M$ & WOR & WR & WPR\\
\toprule
$2$ & 0.56 (2)  & {\chodoiikanjinosize \textbf{0.25}} (1)  & 0.62 (3)  \\
$3$ & 0.25 (2)  & {\chodoiikanjinosize \textbf{0.06}} (1)  & 0.44 (3)  \\
$4$ & {\chodoiikanjinosize \textbf{0.06}} (1)  & 0.31 (3)  & {\chodoiikanjinosize \textbf{0.06}} (1)  \\
$5$ & {\chodoiikanjinosize \textbf{0.06}} (1)  & 0.12 (2)  & 0.31 (3)  \\
\midrule
\end{tabular}
}
\end{table}

\begin{table}[t]
\centering
  \caption{\small 
Overall performance of MOEA/D-DE with the two mutation strategies on the 16 problems.
The APS value at the final iteration is shown in the tables (lower is better).
The numbers in parentheses indicate the ranks of the three configurations based on their APS values.
}
  \label{tab:results_mutation_strategies}
\begin{tabular}{ccccccccc}
\midrule
$M$ & current/1 & rand/1\\
\toprule
$2$ & {\chodoiikanjinosize \textbf{0.12}} (1)  & 0.62 (2)  \\
$3$ & {\chodoiikanjinosize \textbf{0.00}} (1)  & 0.50 (2)  \\
$4$ & {\chodoiikanjinosize \textbf{0.06}} (1)  & 0.38 (2)  \\
$5$ & {\chodoiikanjinosize \textbf{0.06}} (1)  & 0.19 (2)  \\
\midrule
\end{tabular}
\end{table}

\begin{table*}[t]
\centering
  \caption{\small 
Overall performance of MOEA/D-DE with the five bound-handling methods on the 16 problems.
The APS value at the final iteration is shown in the tables (lower is better).
The numbers in parentheses indicate the ranks of the five configurations.
}
  \label{tab:results_bchs}
\begin{tabular}{ccccccccc}
\midrule
$M$ & replacement & reinitialization & reflection & r-reflection & resampling\\
\toprule
$2$ & {\chodoiikanjinosize \textbf{0.69}} (1)  & 2.12 (5)  & 1.38 (3)  & 1.19 (2)  & 1.69 (4)  \\
$3$ & {\chodoiikanjinosize \textbf{0.62}} (1)  & 1.56 (3)  & 2.31 (5)  & 1.88 (4)  & 0.81 (2)  \\
$4$ & {\chodoiikanjinosize \textbf{0.62}} (1)  & 1.69 (3)  & 2.12 (5)  & 1.75 (4)  & 1.06 (2)  \\
$5$ & {\chodoiikanjinosize \textbf{0.81}} (1)  & 1.62 (3)  & 2.31 (5)  & 1.94 (4)  & 1.19 (2)  \\
\midrule
\end{tabular}
\end{table*}

\section{Experimental results}
\label{sec:experimental_results}

This section describes results of MOEA/D-DE with various configurations of the DE mutation operator.
First, Subsection \ref{sec:results_isms} analyzes the effect of the index selection method in MOEA/D-DE.
Next, Subsection \ref{sec:results_mutation_strategies} investigates the influence of the mutation strategy on MOEA/D-DE.
Subsection \ref{sec:results_bchs} presents a comparison of the five bound-handling methods.
Subsection \ref{sec:results_all_configs} examines which configuration of the DE mutation operator is the most suitable for MOEA/D-DE.
We examine the effectiveness of a total of 30 configurations of three index selection methods (WOR, WR, and WPR), two mutation strategies (rand/1 and current/1), and five bound-handling methods (resampling, replacement, reinitialization, reflection, and r-reflection).
Finally, Subsection \ref{sec:results_existing_codes} compares the eight existing configurations (\#A, ..., \#H) shown in Table  \ref{tab:moead_de_source_code}.

\subsection{The impact of the index selection method on the performance of MOEA/D-DE}
\label{sec:results_isms}

Table \ref{tab:results_psm} shows the overall performance of MOEA/D-DE with the three index selection methods (the WOR, WR, and WPR methods) on the 16 problems with $M \in \{2, 3, 4, 5\}$.
Tables \ref{tab:results_psm}(a) and (b) show the results of MOEA/D-DE with the current/1 and rand/1 strategies, respectively.
We do not present the comparison results of each problem here due to space constraints, but they can be found in Tables S.1 and S.2 in the supplementary file of this paper.



Table \ref{tab:results_psm}(a) shows that the best performance of MOEA /D-DE with the current/1 strategy is obtained by using the WPR selection method for $M=2$. 
Although MOEA/D-DE with the WOR and WR methods achieve the same APS value for $M=4$, the WR method is the best selection method for $M \in \{3, 4, 5\}$.

%

Table \ref{tab:results_psm}(b) indicates that the WR selection method is the most suitable for MOEA/D-DE using the rand/1 strategy  for $M \in \{2, 3\}$.
In contrast to the results for $M \in \{2, 3\}$, MOEA/D-DE with the WR method performs poorly for $M \in \{4, 5\}$.
Although MOEA/D-DE with the WOR and WPR methods show the same APS value for $M=4$, the WOR method is the best selection method for $M = 5$.

In summary, our results indicate that the choice of index selection method significantly affects the performance of MOEA/D-DE.
As described above, the best index selection method depends on the type of mutation strategies used in MOEA/D-DE and on the number of objectives $M$.
In addition, the most suitable index selection method also depends on the type of problems (Tables S.1 and S.2).
However, on most problems, the WR and WOR methods are likely to be suitable for the current/1 and rand/1 mutation strategies, respectively.



\subsection{The rand/1 and current/1 mutation strategies, which is better for MOEA/D-DE?}
\label{sec:results_mutation_strategies}

Table \ref{tab:results_mutation_strategies} shows the overall performance of MOEA/D-DE with the current/1 and rand/1 mutation strategies on the 16 problems with $M \in \{2, 3, 4, 5\}$.
Based on the results in Subsection \ref{sec:results_isms}, the WR and WOR methods were used for the current/1 and rand/1 mutation strategies, respectively.
Results on each problem are presented in Table S.3 in the supplementary file.

Table \ref{tab:results_mutation_strategies} shows that the current/1 strategy is more suitable for MOEA/D-DE than the rand/1 strategy for all $M$.
As discussed in \cite{GongWCY17}, the mutant vector $\vector{v}^i$ generated by the current/1 strategy is generally close to the target individual $\vector{x}^i$, and the neighborhood region of $\vector{x}^i$ is efficiently exploited, which is the reason why MOEA/D-DE with the current/1 strategy performs well.
Although the rand/1 mutation strategy is incorporated into some packages, as reviewed in Subsection \ref{sec:de_mutation_operator}, Table \ref{tab:results_mutation_strategies} suggests that the current/1 strategy is more appropriate for MOEA/D-DE.

\begin{table*}[t]
\renewcommand{\arraystretch}{0.5}
\centering
  \caption{\small 
Overall performance of MOEA/D-DE with all 30 configurations on the 16 problems with $M \in \{2, 3, 4, 5\}$.
The rank of each configuration based on its APS values at the final iteration is shown for each $M$.
The average rank for all $M$ is also reported. 
}
{\footnotesize
  \label{tab:results_all_30}
\scalebox{1}[1]{
\begin{tabular}{cccccccccc}
\midrule
\shortstack{Bound-\\handling\\methods} & \raisebox{0.75em}{\shortstack{Mutation\\strategies}} & \shortstack{Index\\selection\\methods} & \raisebox{0.75em}{\shortstack{Configuration\\number \#}} & \raisebox{1em}{$M = 2$} & \raisebox{1em}{$M = 3$} & \raisebox{1em}{$M = 4$} & \raisebox{1em}{$M = 5$} & \raisebox{1em}{Avg. rank}\\
\toprule
 &  & WOR & \#F & 5 & 3 & 1 & 4 & 3.25\\  \cmidrule(l{.75em}r{.75em}){3-9}
 & current/1 & WR & & 1 & 1 & 3 & 1 & 1.5\\  \cmidrule(l{.75em}r{.75em}){3-9}
\multirow{3}{*}{replacement} &  & WPR & \#B & 2 & 2 & 2 & 2 & 2.0\\ \cmidrule(l{.75em}r{.75em}){2-9}
 &  & WOR & \#C & 11 & 6 & 5 & 2 & 6.0\\  \cmidrule(l{.75em}r{.75em}){3-9}
 & rand/1 & WR & & 4 & 4 & 6 & 4 & 4.5\\  \cmidrule(l{.75em}r{.75em}){3-9}
 &  & WPR & \#D & 12 & 7 & 4 & 6 & 7.25\\  \midrule
 &  & WOR & & 20 & 20 & 22 & 16 & 19.5\\  \cmidrule(l{.75em}r{.75em}){3-9}
 & current/1 & WR &  \#A & 16 & 8 & 10 & 11 & 11.25\\  \cmidrule(l{.75em}r{.75em}){3-9}
\multirow{3}{*}{reinitialization} &  & WPR & & 15 & 16 & 18 & 14 & 15.75\\ \cmidrule(l{.75em}r{.75em}){2-9}
 &  & WOR & & 29 & 26 & 27 & 19 & 25.25\\  \cmidrule(l{.75em}r{.75em}){3-9}
 & rand/1 & WR & & 22 & 14 & 16 & 15 & 16.75\\  \cmidrule(l{.75em}r{.75em}){3-9}
 &  & WPR & & 30 & 27 & 28 & 22 & 26.75\\  \midrule
 &  & WOR & & 16 & 21 & 22 & 23 & 20.5\\  \cmidrule(l{.75em}r{.75em}){3-9}
 & current/1 & WR & & 8 & 12 & 13 & 20 & 13.25\\  \cmidrule(l{.75em}r{.75em}){3-9}
\multirow{3}{*}{reflection} &  & WPR & & 7 & 18 & 21 & 21 & 16.75\\ \cmidrule(l{.75em}r{.75em}){2-9}
 &  & WOR & & 28 & 29 & 29 & 30 & 29.0\\  \cmidrule(l{.75em}r{.75em}){3-9}
 & rand/1 & WR & & 19 & 22 & 24 & 24 & 22.25\\  \cmidrule(l{.75em}r{.75em}){3-9}
 &  & WPR & & 25 & 30 & 30 & 29 & 28.5\\  \midrule
 &  & WOR & \#E & 18 & 17 & 14 & 28 & 19.25\\  \cmidrule(l{.75em}r{.75em}){3-9}
 & current/1 & WR & & 3 & 11 & 11 & 17 & 10.5\\  \cmidrule(l{.75em}r{.75em}){3-9}
\multirow{3}{*}{r-reflection} & & WPR & \#H & 6 & 15 & 19 & 26 & 16.5\\ \cmidrule(l{.75em}r{.75em}){2-9}
 &  & WOR & \#G & 23 & 22 & 26 & 25 & 24.0\\  \cmidrule(l{.75em}r{.75em}){3-9}
 & rand/1 & WR & & 13 & 19 & 20 & 18 & 17.5\\  \cmidrule(l{.75em}r{.75em}){3-9}
 &  & WPR & & 24 & 24 & 25 & 27 & 25.0\\  \midrule
 &  & WOR & & 14 & 10 & 8 & 8 & 10.0\\  \cmidrule(l{.75em}r{.75em}){3-9}
 & current/1 & WR & & 10 & 5 & 7 & 7 & 7.25\\  \cmidrule(l{.75em}r{.75em}){3-9}
\multirow{3}{*}{resampling} &  & WPR & & 9 & 9 & 9 & 10 & 9.25\\ \cmidrule(l{.75em}r{.75em}){2-9}
 &  & WOR & & 26 & 28 & 14 & 12 & 20.0\\  \cmidrule(l{.75em}r{.75em}){3-9}
 & rand/1 & WR & & 21 & 13 & 12 & 9 & 13.75\\  \cmidrule(l{.75em}r{.75em}){3-9}
 &  & WPR & & 27 & 25 & 17 & 13 & 20.5\\  \midrule
\end{tabular}
}
}
\end{table*}

 \subsection{The influence of the bound-handling method on the performance of MOEA/D-DE}
 \label{sec:results_bchs}

Table \ref{tab:results_bchs} shows the overall performance of MOEA/D-DE with the five bound-handling methods on the 16 problems.
Results on each MOP can be found in Table S.4 in the supplementary file.

For all $M$, the best results are obtained by using the replacement method, followed by the resampling method.
For $M \geq 3$, MOEA/D-DE with the reflection method achieves the worst APS values.
Although the r-reflection method is used in some implementations of MOEA/D-DE, as reviewed in Subsection \ref{sec:bchs}, its APS value is poor for $M \geq 3$.
In summary, our results show that the replacement method is the most suitable bound-handling method of the DE mutation operator in MOEA/D-DE.

The results reported in \cite{ArabasSW10} show that the resampling method is the most appropriate for the basic DE \cite{StornP97} for single-objective continuous optimization.
In contrast to the results presented in \cite{ArabasSW10}, MOEA/D-DE with the resampling method performs the second best on MOPs.
Thus, the most suitable bound-handling method for the DE mutation operator depends on the problem domain.

\begin{table*}[t]
\renewcommand{\arraystretch}{0.5}
\centering
\caption{\small
Performance of MOEA/D-DE with all 30 configurations on the four problem sets (unimodal, multimodal, separable, and nonseparable problems).
The average rank values for all $M \in \{2, 3, 4, 5\}$ are shown.
}
{\footnotesize
  \label{tab:results_each_problem_type}
\scalebox{1}[1]{
\begin{tabular}{ccccccccc}
\midrule
\shortstack{Bound-\\handling\\methods} & \raisebox{0.75em}{\shortstack{Mutation\\strategies}} & \shortstack{Index\\selection\\methods}  & \raisebox{0.75em}{\shortstack{Configuration\\number \#}} & \raisebox{1em}{Unimodal} & \raisebox{1em}{Multimodal} & \raisebox{1em}{Separable} & \raisebox{1em}{Nonseparable}\\
\toprule
 &  & WOR & \#F & 3.25 & 7.0 & 3.5 & 9.5\\  \cmidrule(l{.75em}r{.75em}){3-8}
 & current/1 & WR & &  1.25 & 4.25 & 1.0 & 12.5\\  \cmidrule(l{.75em}r{.75em}){3-8}
\multirow{3}{*}{replacement} &  & WPR & \#B & 2.25 & 6.5 & 2.25 & 9.5\\ \cmidrule(l{.75em}r{.75em}){2-8}
 &  & WOR  & \#C & 5.5 & 10.0 & 5.75 & 13.25\\  \cmidrule(l{.75em}r{.75em}){3-8}
 & rand/1 & WR & & 3.5 & 8.0 & 3.75 & 17.0\\  \cmidrule(l{.75em}r{.75em}){3-8}
 &  & WPR  & \#D & 6.25 & 10.5 & 5.75 & 13.25\\  \midrule
 &  & WOR  & & 22.5 & 10.25 & 24.0 & 8.25\\  \cmidrule(l{.75em}r{.75em}){3-8}
 & current/1 & WR  & \#A & 13.0 & 5.75 & 13.5 & 2.5\\  \cmidrule(l{.75em}r{.75em}){3-8}
\multirow{3}{*}{reinitialization} &  & WPR & & 16.5 & 8.0 & 20.25 & 3.0\\ \cmidrule(l{.75em}r{.75em}){2-8}
 &  & WOR & & 27.25 & 16.5 & 27.5 & 18.5\\  \cmidrule(l{.75em}r{.75em}){3-8}
 & rand/1 & WR & & 18.75 & 9.25 & 18.75 & 11.75\\  \cmidrule(l{.75em}r{.75em}){3-8}
 &  & WPR & & 28.75 & 16.75 & 29.0 & 16.75\\  \midrule
 &  & WOR &  & 20.25 & 21.0 & 20.75 & 16.5\\  \cmidrule(l{.75em}r{.75em}){3-8}
 & current/1 & WR & & 12.5 & 20.5 & 11.0 & 20.75\\  \cmidrule(l{.75em}r{.75em}){3-8}
\multirow{3}{*}{reflection} &  & WPR & & 15.5 & 17.75 & 16.25 & 16.5\\ \cmidrule(l{.75em}r{.75em}){2-8}
 &  & WOR & & 27.25 & 26.0 & 28.0 & 24.25\\  \cmidrule(l{.75em}r{.75em}){3-8}
 & rand/1 & WR & & 20.25 & 24.5 & 18.5 & 25.25\\  \cmidrule(l{.75em}r{.75em}){3-8}
 &  & WPR & & 28.0 & 26.75 & 27.5 & 25.25\\  \midrule
 &  & WOR & \#E & 18.0 & 21.75 & 18.5 & 17.5\\  \cmidrule(l{.75em}r{.75em}){3-8}
 & current/1 & WR & & 10.0 & 16.5 & 8.5 & 15.75\\  \cmidrule(l{.75em}r{.75em}){3-8}
\multirow{3}{*}{r-reflection} &  & WPR & \#H & 15.5 & 17.5 & 14.75 & 17.25\\ \cmidrule(l{.75em}r{.75em}){2-8}
 &  & WOR & \#G & 22.75 & 23.0 & 23.75 & 21.75\\  \cmidrule(l{.75em}r{.75em}){3-8}
 & rand/1 & WR & & 15.0 & 23.0 & 14.0 & 20.5\\  \cmidrule(l{.75em}r{.75em}){3-8}
 &  & WPR & & 23.5 & 25.25 & 24.25 & 22.5\\  \midrule
 &  & WOR & & 9.75 & 12.5 & 10.0 & 8.25\\  \cmidrule(l{.75em}r{.75em}){3-8}
 & current/1 & WR & & 8.0 & 6.5 & 7.5 & 7.5\\  \cmidrule(l{.75em}r{.75em}){3-8}
\multirow{3}{*}{resampling} &  & WPR & & 9.5 & 10.25 & 10.0 & 8.5\\ \cmidrule(l{.75em}r{.75em}){2-8}
 &  & WOR & & 19.75 & 19.25 & 19.75 & 17.5\\  \cmidrule(l{.75em}r{.75em}){3-8}
 & rand/1 & WR & & 13.75 & 17.0 & 13.25 & 17.0\\  \cmidrule(l{.75em}r{.75em}){3-8}
 &  & WPR & & 21.5 & 18.0 & 21.0 & 17.25\\  \midrule
\end{tabular}
\color{black}
}
}
\end{table*}

 \subsection{Comparison of all 30 configurations}
 \label{sec:results_all_configs}

We separately examined the effect of the three components of the DE mutation operator in Subsections \ref{sec:results_isms}, \ref{sec:results_mutation_strategies}, and \ref{sec:results_bchs}.
The results in the three subsections show that the WR method, the current/1 strategy, and the replacement method are suitable in MOEA/D-DE.
However, it is still unclear whether MOEA/D-DE with a combination of the three well-performing elements works well.
To determine whether it performs well, this section presents comparisons of MOEA/D-DE with a total of 30 configurations (the three index selection methods, the two mutation strategies, and the five bound-handling methods) of the DE mutation operator.

First, we show the results on the 16 test problems in Subsection \ref{sec:results_all_configs_16}.
Then, we report the performance of the 30 configurations on each problem type in Subsection \ref{sec:results_all_configs_each}.


 \subsubsection{Results on the 16 test problems}
 \label{sec:results_all_configs_16}
 
Table \ref{tab:results_all_30} shows the overall performance of MOEA/D-DE with all 30 configurations on the 16 problems.
Below, $(b, m, s)$-MOEA/D-DE denotes an MOEA/D-DE using $b$, $m$, and $s$, where $b$, $m$, and $s$ represent a bound-handling method, a mutation strategy, and an index selection method, respectively.
%


Table \ref{tab:results_all_30} indicates that the performance rank of MOEA /D-DE significantly depends on the configuration of the DE mutation operator.
For example, (replacement, current/1, WPR)-MOEA/D-DE works well for any number of objectives, and its average rank is $2.0$.
In contrast, the worst configuration is (reflection, rand /1, WOR)-MOEA/D-DE, whose average rank is $29.0$.
Although this configuration is frequently incorporated into DE for single-objective continuous optimization (e.g., \cite{RonkkonenKP05,WangCZ11}), our results show that it is unsuitable for MOEA /D-DE.
As shown in Table \ref{tab:results_all_30}, the best configuration depends on the number of objectives $M$.
According to the average rank for all $M$, the best performance is obtained by (replacement, current/1, WR)-MOEA/D-DE.
This configuration consists of the three well-performing components, as shown in Subsections \ref{sec:results_isms} -- \ref{sec:results_bchs}.


%
%

%

 \subsubsection{Results on each problem type}
 \label{sec:results_all_configs_each}
 
Next, we discuss the performance of the 30 configurations on each problem type.
According to Table \ref{tab:wfg_properties}, we classified the 16 test problems into the following four groups:
\begin{description}
  \item[Unimodal problems] DTLZ2, DTLZ4, DTLZ5,  DTLZ6, WFG1, WFG3, WFG5, WFG6, WFG7, and WFG8
  \item[Multimodal problems] DTLZ1, DTLZ3, DTLZ7, WFG2, WFG4, and WFG9
  \item[Separable problems] DTLZ1, DTLZ2, DTLZ3, DTLZ4, DTLZ5,  DTLZ6,  DTLZ7, WFG1, WFG4, WFG5, and WFG7
  \item[Nonseparable problems] WFG2, WFG3, WFG6, WFG8, and WFG9
\end{description}

Table \ref{tab:results_each_problem_type} shows results of the 30 configurations on the four problem types.
Each value in Table \ref{tab:results_each_problem_type} is the average rank value of each configuration over all problems ($M \in \{2, 3, 4, 5\}$) in each problem type.
%
On the one hand, Table \ref{tab:results_each_problem_type} indicates that the performance rank of each configuration for the unimodal, multimodal, and separable problem sets is almost consistent with that for all 16 test problems in Table \ref{tab:results_all_30}.
The best average rank value is obtained by (replacement, current/1, WR)-MOEA/D-DE.
In addition, (replacement, current/1, WPR)-MOEA/D-DE performs well.
On the other hand, the results for the nonseparable problem set show that the reinitialization repair method is suitable for MOEA/D-DE.
(reinitialization, current/1, WR)-MOEA/D-DE and (reinitialization, current/1, WPR)-MOEA/D-DE have good performance according to the average rank values.
This result may be because MOEA/D-DE needs to keep the diversity of the population to find good solutions on nonseparable problems.
Since the reinitialization method randomly generates an element of a child that is out of the bound, it can introduce diversity in the population.
Although the best configuration also depends on the problem type, (replacement, current/1, WR)-MOEA/D-DE works well on a wide variety of problems.

\subsection{Comparison between the eight existing configurations}
\label{sec:results_existing_codes}

As reviewed in Section \ref{sec:review}, the configuration of the DE mutation operator differs depending on the source code. 
According to Table \ref{tab:moead_de_source_code}, eight configurations (\#A, ..., \#H) exist.
Our results in Subsections \ref{sec:results_isms}--\ref{sec:results_all_configs} also show that the performance of MOEA/D-DE significantly depends on components of the mutation.
Here, we discuss the performance rank of MOEA/D-DE with the eight existing configurations. 

Table \ref{tab:results_all_30} shows that configuration \#B has the best average rank value among the eight configurations.
Configuration \#F is the second-best performer.
Although the average rank value of configuration \#A is not good for all 16 problems, it is best for the nonseparable problem set, as discussed in Subsection \ref{sec:results_all_configs_each} (see Table \ref{tab:results_each_problem_type}).
In contrast, configurations \#E, \#G, and \#H show poor performance.
In particular, configuration \#G performs the worst among the eight configurations (\#A, ..., \#H).

The above results indicate that an incorrect conclusion could be obtained depending on the source code of MOEA/D-DE used.
For example, let us consider the experiment to reproduce the results of MOEA/D-DE in the original paper \cite{LiZ09}.
As shown in Table \ref{tab:moead_de_source_code}, configuration \#B is used in the original C++ code of MOEA/D-DE.
However, configuration \#G is incorporated into the PaGMO library.
If one uses the PaGMO implementation of MOEA/D-DE (\#G) rather than its original C++ source code (\#B), she/he is likely to obtain results that are worse than those reported in \cite{LiZ09}.
That is, the performance of MOEA/D-DE is underestimated in this case.
To avoid such an undesirable comparison, the configuration of the DE mutation operator should receive careful attention.

\section{Conclusion}
\label{sec:conclusion}



We have reviewed the existing configurations of the DE mutation in MOEA/D-DE (Section \ref{sec:review}) and examined the influence of each component on the performance of MOEA/D-DE (Section \ref{sec:experimental_results}).
The main contributions of this paper can be summarized as follows:
\begin{itemize}
\item The DE mutation consists of three components (the mutation strategy, the index selection method, and the bound-handling method). Our review reveals that the configuration of the DE mutation operator in MOEA/D-DE differs depending on the source code.
We also explained why different configurations of the DE mutation operator are implemented in the source code of MOEA/D-DE.
\item Although implementations of the DE mutation operator in MOEA/D-DE have not received much attention, our results shows that the performance of MOEA/D-DE significantly depends on the configuration of the DE mutation operator.
Thus, it is necessary to carefully select the configuration of the DE mutation operator to maximize the performance of MOEA/D-DE.
Our results show that the combination of the current/1 strategy, the WR method (or the WPR method), and the replacement method is most suitable for MOEA/D-DE.
\end{itemize}

As we discussed in Section \ref{sec:experimental_results}, some of our results on MOPs are inconsistent with previous studies on single-objective optimization problems.
For example, while a combination of the WOR method, the rand/1 strategy, and the reflection method is frequently used in DE algorithms for single-objective optimization, our results indicates that it is inappropriate for MOEA/D-DE.
As far as we know, such observations have not been reported in the literature.
One future research topic is to investigate why the best configuration of the DE mutation operator depends on the problem domain.

Another future research topic is to analyze the impact of configurations of the DE mutation operator on the performance of other DE-based MOEAs (e.g., DEMO \cite{RobicF05} and GDE3\cite{KukkonenL05}).
We used the DTLZ and WFG problems to examine the performance of MOEA/D-DE. 
Comparison of various configurations of the DE mutation in MOEA/D-DE on other test problems (e.g., the CEC2017 problems \cite{ChengLTZYJY17}) remains as future work.

 \section*{Acknowledgement}


This work was supported by the Program for Guangdong Introducing Innovative and Entrepreneurial Teams (Grant No. 2017ZT07X386), Shenzhen Peacock Plan (Grant No. KQTD2016112514355531), the Science and Technology Innovation Committee Foundation of Shenzhen (Grant No. ZDSYS201703031748284), the Program for University Key Laboratory of Guangdong Province (Grant No. 2017KSYS008), and National Natural Science Foundation of China (Grant No. 61876075).

 \section*{Compliance with ethical standards}


 \section*{Conflict of interest} Ryoji Tanabe and Hisao Ishibuchi declare that they have no conflict of interest.

 \section*{Ethical approval} This article does not contain any studies with human participants performed by any of the authors.

\bibliographystyle{plain}
\bibliography{reference}

\end{document}